\documentclass[mph,amsmath,amssymb,mph,reprint,graphicx,nofootinbib]{revtex4-1} 
\usepackage[pdftex]{graphicx}
\usepackage[colorlinks=true,linkcolor=blue,citecolor=blue]{hyperref}
\DeclareGraphicsExtensions{.pdf,.jpeg,.png}
\usepackage{todonotes}

\usepackage{url}
\usepackage{amsmath,amsfonts,amssymb,bm}
\usepackage{graphicx}
\usepackage{color}
\usepackage{tabularx}
\usepackage[mathlines]{lineno}
\draft 
\begin{document}


\title{AnatomyNet: Deep Learning for Fast  and Fully Automated Whole-volume Segmentation of  Head and Neck Anatomy} 



\author{Wentao Zhu} 
\affiliation{Department of Computer Science, University of California, Irvine} 
\author{Yufang Huang}
\affiliation{Lenovo Research, Beijing, China}
\author{Liang Zeng}
\affiliation{Deepvoxel Inc, Shanghai, China}
\author{Xuming Chen and Yong Liu}
\affiliation{Department of Radiation Oncology, Shanghai General Hospital, Shanghai Jiao Tong University School of Medicine, Shanghai, China}
\author{Zhen Qian, Nan Du and Wei Fan}
\affiliation{Tencent Medical AI Lab, Palo Alto, CA }
\author{Xiaohui Xie} 
\affiliation{Department of Computer Science, University of California, Irvine, CA, USA} 
    \email[]{xhx@ics.uci.edu}
\thanks{Correspondence author}


\date{\today}

\begin{abstract}
\textbf{Purpose:} Radiation therapy (RT) is a common treatment option for head and neck (HaN) cancer. An important step involved in RT planning is the delineation of organs-at-risks (OARs) based on HaN computed tomography (CT). However, manually delineating OARs is time-consuming as each slice of CT images needs to be individually examined and a typical CT consists of hundreds of slices. Automating OARs segmentation has the benefit of both reducing the time and improving the quality of RT planning. Existing anatomy auto-segmentation algorithms use primarily atlas-based methods, which require sophisticated atlas creation and cannot adequately account for anatomy variations among patients. In this work, we propose an end-to-end, atlas-free 3D convolutional deep learning framework for fast and fully automated whole-volume HaN anatomy segmentation. 

\textbf{Methods:}
Our deep learning model, called AnatomyNet, segments OARs from head and neck CT images in an end-to-end fashion, receiving whole-volume HaN CT images as input and generating masks of all OARs of interest in one shot. AnatomyNet is built upon the popular 3D U-net architecture, but extends it in three important ways: 1) a new encoding scheme to allow auto-segmentation on whole-volume CT images instead of local patches or subsets of slices, 2) incorporating 3D squeeze-and-excitation residual blocks in encoding layers for better feature representation, and 3) a new loss function combining Dice scores and focal loss to facilitate the training of the neural model. These features are designed to address two main challenges in deep-learning-based HaN segmentation: a)  segmenting small anatomies (i.e., optic chiasm and optic nerves) occupying only a few slices, and b) training with inconsistent data annotations with missing ground truth for some anatomical structures.

\textbf{Results:}
We collected 261 HaN CT images to train AnatomyNet, and used MICCAI Head and Neck Auto Segmentation Challenge 2015 as a benchmark dataset to evaluate the performance of AnatomyNet. The objective is to segment nine anatomies: brain stem, chiasm, mandible, optic nerve left, optic nerve right, parotid gland left, parotid gland right, submandibular gland left, and submandibular gland right. Compared to previous state-of-the-art results from the MICCAI 2015 competition, AnatomyNet increases Dice similarity coefficient by 3.3\% on average. AnatomyNet takes about 0.12 seconds to fully segment  a head and neck CT image of  dimension  $178 \times 302 \times 225$, significantly faster than previous methods. In addition, the model is able to process  whole-volume CT images and delineate all OARs in one pass, requiring little pre- or post-processing.

\textbf{Conclusion:}
Deep learning models offer a feasible solution to the problem of delineating OARs from CT images. We demonstrate that our proposed model can improve segmentation accuracy and simplify the auto-segmentation pipeline. With this method, it is possible to delineate OARs of a head and neck CT within a fraction of a second.

\end{abstract}


\maketitle 

\section{INTRODUCTION}
\label{sec:intro}  
Head and neck cancer is one of the most common cancers around the world \cite{torre2015global}. Radiation therapy is the primary method for treating patients with head and neck cancers. The planning of the radiation therapy relies on accurate organs-at-risks (OARs) segmentation \cite{han2008atlas}, which is usually undertaken by radiation therapists with laborious manual delineation. Computational tools that automatically segment the anatomical regions can greatly alleviate doctors' manual efforts if these tools can delineate anatomical regions accurately with a reasonable amount of time \cite{sharp2014vision}. 

There is a vast body of literature on automatically segmenting anatomical structures from CT or MRI images. Here we focus on reviewing literature related to head and neck (HaN) CT anatomy segmentation. Traditional anatomical segmentation methods use primarily atlas-based methods, producing segmentations by aligning new images to a fixed set of manually labelled exemplars \cite{raudaschl2017evaluation}. Atlas-based segmentation methods typically undergo a few steps, including preprocessing, atlas creation, image registration, and label fusion. As a consequence, their performances can be affected by various factors involved in each of these steps, such as methods for creating atlas \cite{han2008atlas,voet2011does,isambert2008evaluation,fritscher2014automatic,commowick2008atlas,sims2009pre,fortunati2013tissue,verhaart2014relevance,wachinger2015contour}, methods for label fusion \cite{duc2015validation,duc2015validation,fortunati2015automatic}, and  methods for registration \cite{zhang2007automatic,chen2010combining,han2008atlas,duc2015validation,fritscher2014automatic,fortunati2013tissue,wachinger2015contour,qazi2011auto,leavens2008validation}.  Although atlas-based methods are still very popular and by far the most widely used methods in anatomy segmentation, their main limitation is the difficulty to handle anatomy variations among patients because they use a fixed set of atlas. In addition, it is computationally intensive and can take many minutes to complete one registration task even with most efficient implementations \cite{xu2018use}.



Instead of aligning images to a fixed set of exemplars, learning-based methods trained to directly segment OARs without resorting to reference exemplars have also been tried \cite{tam2018automated,wu2018auto,tong2018hierarchical,pednekar2018image,wang2018hierarchical}. However, most of the learning-based methods require laborious preprocessing steps, and/or hand-crafted image features. As a result, their performances tend to be less robust than registration-based methods.

Recently, deep convolutional models have shown great success for biomedical image segmentation \cite{ronneberger2015u}, and have been introduced to the field of HaN anatomy segmentation \cite{fritscher2016deep,ibragimov2017segmentation,ren2018interleaved,hansch2018comparison}. However, the existing HaN-related deep-learning-based methods either use sliding windows working on patches that cannot capture global features, or rely on atlas registration to obtain highly accurate small regions of interest in the preprocessing. What is more appealing are models that receive the whole-volume image as input without heavy-duty preprocessing, and then directly output the segmentations of all interested anatomies. 

In this work, we study the feasibility and performance of constructing and training a deep neural net model that jointly segment all OARs in a fully end-to-end fashion, receiving  raw whole-volume HaN CT images as input and generating the masks of all OARs in one shot. The success of such a system can improve the current performance of automated anatomy segmentation by simplifying the entire computational pipeline, cutting computational cost and improving segmentation accuracy.


There are, however, a number of obstacles that need to overcome in order to make such a deep convolutional neural net based system successful. First, in designing network architectures, we ought to keep the maximum capacity of GPU memories in mind. Since whole-volume images are used as input, each image feature map will be 3D, limiting the size and number of feature maps at each layer of the neural net due to memory constraints.  Second, OARs contain organs/regions of variable sizes, including some OARs with very small sizes. Accurately segmenting these small-volumed structures is always a challenge. Third, existing datasets of HaN CT images contain data collected from various sources with non-standardized annotations. In particular, many images in the training data contain annotations of only a subset of OARs. How to effectively handle missing annotations needs to be addressed in the design of the training algorithms. 

Here we propose a deep learning based framework, called AnatomyNet, to segment OARs using a single network, trained end-to-end. The network receives whole-volume CT images as input, and outputs the segmented masks of all OARs. Our method requires minimal pre- and post-processing, and utilizes features from all slices to segment anatomical regions. We overcome the three major obstacles outlined above through designing a novel network architecture and utilizing novel loss functions for training the network. 

More specifically, our major contributions include the following. First, we extend the standard U-Net model for 3D HaN image segmentation by incorporating a new feature extraction component, based on squeeze-and-excitation (SE) residual blocks \cite{hu2017squeeze}. Second, we propose a new loss function for better segmenting small-volumed structures. Small volume segmentation suffers from the imbalanced data problem, where the number of voxels inside the small region is much smaller than those outside, leading to the difficulty of training. New classes of loss functions have been proposed to address this issue, including Tversky loss \cite{salehi2017tversky}, generalized Dice coefficients \cite{crum2006generalized,sudre2017generalised}, focal loss \cite{lin2017focal}, sparsity label assignment deep multi-instance learning \cite{zhu2017deep}, and exponential logarithm loss. However, we found none of these solutions alone was adequate to solve the extremely data imbalanced problem (1/100,000) we face in segmenting small OARs, such as optic nerves and chiasm, from HaN images. We propose a new loss based on the combination of Dice scores and focal losses, and empirically show that it leads to better results than other losses. Finally, to tackle the missing annotation problem, we train the AnatomyNet with masked and weighted loss function to account for missing data and to balance the contributions of the losses originating from different OARs.  


To train and evaluate the performance of AnatomyNet, we curated a dataset of 261 head and neck CT images from a number of publicly available sources. We carried out systematic experimental analyses on various components of the network, and demonstrated their effectiveness by comparing with other published methods.  When benchmarked on the test dataset from the MICCAI 2015 competition on HaN segmentation, the AnatomyNet outperformed the state-of-the-art method by 
\textbf{3.3\%} in terms of  Dice coefficient (DSC), averaged over nine anatomical structures.


The rest of the paper is organized as follows. Section \ref{sec:anatomynet} describes the network structure and SE residual block of AnatomyNet. The designing of the loss function for AnatomyNet is present in Section \ref{sec:extremeimbalance}. How to handle missing annotations is addressed in Section \ref{sec:missingannot}. Section \ref{sec:result} validates the effectiveness of the proposed networks and components. Discussions and limitations are in Section \ref{sec:discus}. We conclude the work in Section \ref{sec:conclu}.

\section{MATERIALS AND METHODS}\label{sec:method}
Next we describe our deep learning model to delineate OARs from head and neck CT images. Our model receives whole-volume HaN CT images of a patient as input and outputs the 3D binary masks of all OARs at once. The dimension of a typical HaN CT is around $178 \times 512 \times 512$, but the sizes can vary across different patients because of image cropping and different settings.  In this work, we focus on segmenting nine OARs most relevant to head and neck cancer radiation therapy - brain stem, chiasm, mandible, optic nerve left, optic nerve right, parotid gland left, parotid gland right, submandibular gland left, and submandibular gland right. Therefore, our model will produce nine 3D binary masks for each whole volume CT. 

\subsection{Data}\label{sec:data}
\vspace{-0.1in}
Before we introduce our model, we first describe the curation of training and testing data. Our data consists of whole-volume CT images together with manually generated binary masks of the nine anatomies described above. There were collected from four publicly available sources: 1) DATASET 1 (38 samples) consists of the training set from the MICCAI Head and Neck Auto Segmentation Challenge 2015 \cite{raudaschl2017evaluation}. 2) DATASET 2 (46 samples) consists of CT images from the Head-Neck Cetuximab collection, downloaded from The Cancer Imaging Archive (TCIA)\footnote{https://wiki.cancerimagingarchive.net/} \cite{clark2013cancer}. 3) DATASET 3 (177 samples) consists of CT images from four different institutions in Qu\'ebec, Canada \cite{vallieres2017radiomics}, also downloaded from TCIA \cite{clark2013cancer}. 4) DATATSET 4 (10 samples) consists of the test set from the MICCAI HaN Segmentation Challenge 2015. We combined the first three datasets and used the aggregated data as our training data, altogether yielding 261 training samples.  DATASET 4 was used as our final evaluation/test dataset so that we can benchmark our performance against published results evaluated on the same dataset.  Each of the training and test samples contains both head and neck images and the corresponding manually delineated OARs.

In generating these datasets, We carried out several data cleaning steps, including 1) mapping annotation names named by different doctors in different hospitals into unified annotation names, 2) finding correspondences between the annotations and the CT images, 3) converting annotations in the radiation therapy format into usable ground truth label mask, and 4) removing chest from CT images to focus on head and neck anatomies.  We have taken care to make sure that the four datasets described above are non-overlapping to avoid any potential pitfall of inflating testing or validation performance. 

\subsection{Network architecture}\label{sec:anatomynet}
\begin{figure*} [ht]
	\begin{center}
		\begin{tabular}{c} 
			\includegraphics[height=7.5cm]{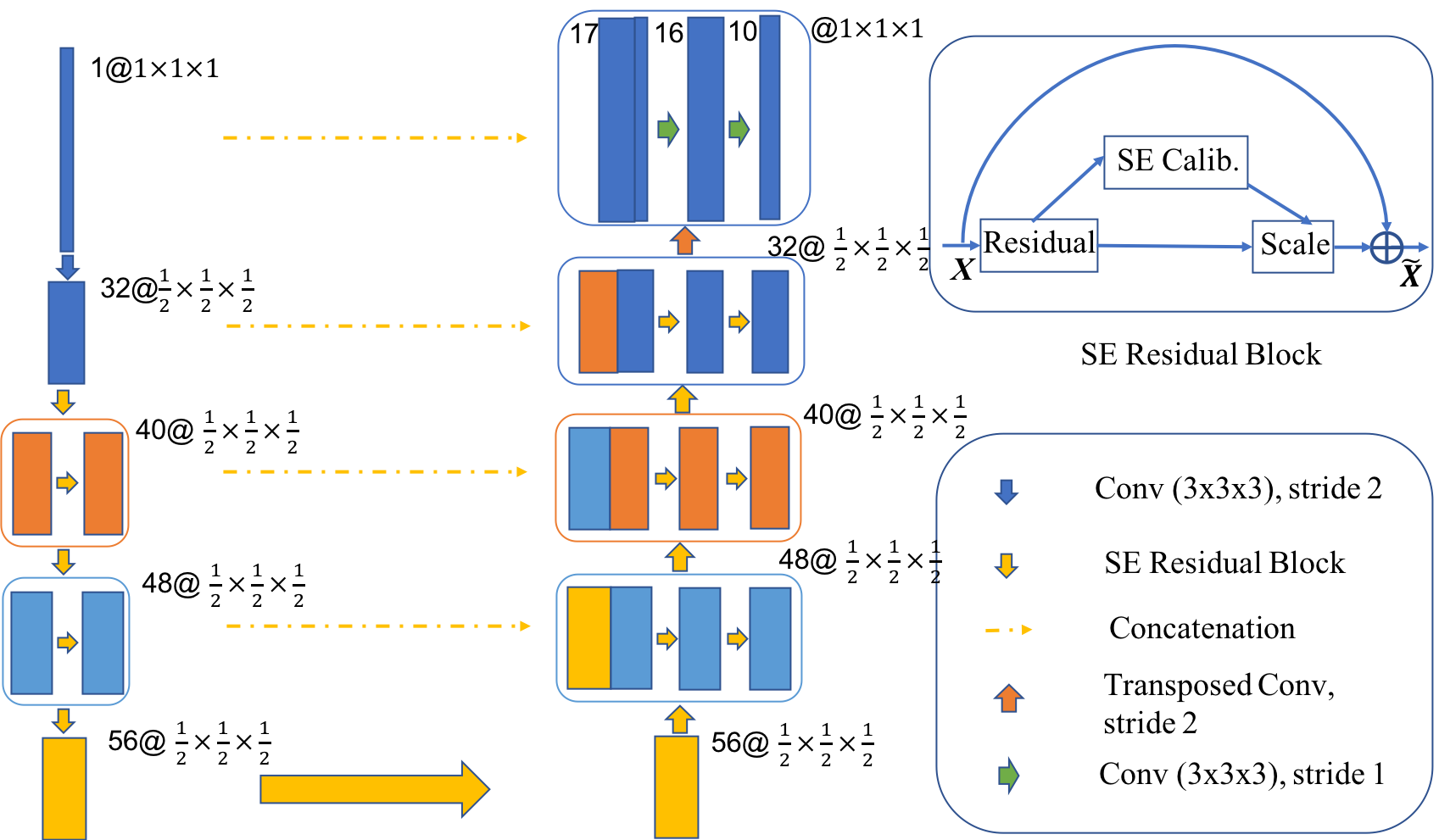}
		\end{tabular}
	\end{center}
\vspace{-0.2in}
	\caption[seunet] 
	{ \label{fig:seunet} 
		The AnatomyNet is a variant of U-Net with only one down-sampling and squeeze-and-excitation (SE) residual building blocks. The number before symbol @ denotes the number of output channels, while the number after the symbol denotes the size of feature map relative to the input. In the decoder, we use concatenated features. Hybrid loss with dice loss and focal loss is employed to force the model to learn not-well-classified voxels. Masked and weighted loss function is used for ground truth with missing annotations and balanced gradient descent respectively. The decoder layers are symmetric with the encoder layers. The SE residual block is illustrated in the upper right corner.}
\end{figure*}

\vspace{-0.1in}
We take advantage of the robust feature learning mechanisms obtained from squeeze-and-excitation (SE) residual blocks \cite{hu2017squeeze}, and incorporate them into a modified U-Net architecture for medical image segmentation. We propose a novel three dimensional U-Net with squeeze-and-excitation (SE) residual blocks and hybrid focal and dice loss for anatomical segmentation as illustrated in Fig. \ref{fig:seunet}. 

The AnatomyNet is a variant of 3D U-Net \cite{ronneberger2015u,zhu2018deeplung,zhu2018deepem}, one of the most commonly used neural net architectures in biomedical image segmentation. The standard U-Net contains multiple down-sampling layers via max-pooling or convolutions with strides over two. Although they are beneficial to learn high-level features for segmenting complex, large anatomies, these down-sampling layers can hurt the segmentation of small anatomies such as optic chiasm, which occupy only a few slices in HaN CT images. We design the AnatomyNet with only one down-sampling layer to account for the trade-off between GPU memory usage and network learning capacity. The down-sampling layer is used in the first encoding block so that the feature maps and gradients in the following layers occupy less GPU memory than other network structures. Inspired by the effectiveness of squeeze-and-excitation residual features on image object classification, we design 3D squeeze-and-excitation (SE) residual blocks in the AnatomyNet for OARs segmentation. The SE residual block adaptively calibrates residual feature maps within each feature channel. 
The 3D SE Residual learning extracts 3D features from CT image directly by extending two-dimensional squeeze, excitation, scale and convolutional functions to three-dimensional functions. It can be formulated as 
\begin{equation}
\label{eq:seres}
\begin{aligned}
{\bm{X}^{r}}&=\bm{F}(\bm{X}) \, , \\
z_{k} &= {\bm{F}}_{sq}({\bm{X}^{r}}_{k}) = \frac{1}{S \times H \times W} \sum_{s=1}^{S} \sum_{h=1}^{H} \sum_{w=1}^{W}{x}^{r}_{k}(s,h,w) \, , \\
\bm{z} &= [z_1, z_2, \cdots, z_{k}, \cdots, z_{K}] \, , \\
\bm{s} &= {\bm{F}}_{ex} (\bm{z}, \bm{W}) = \bm{\sigma}(\bm{W}_2 \bm{G}(\bm{W}_1 \bm{z})) \, , \\
{\tilde{\bm{X}}}_{k} &= {\bm{F}}_{scale}({\bm{X}^{r}}_{k}, s_{k}) = s_{k} {\bm{X}^{r}}_{k} \, , \\
\tilde{\bm{X}} &= [{\tilde{\bm{X}}}_1, {\tilde{\bm{X}}}_2, \cdots, {\tilde{\bm{X}}}_{k}, \cdots, {\tilde{\bm{X}}}_{K}] \, , \\
\bm{Y} &= \bm{G} (\tilde{\bm{X}} + \bm{X}) \, ,
\end{aligned}
\end{equation}
where ${{\bm{X}^{r}}_{k}} \in {\mathbb{R}}^3$ denotes the feature map of one channel from the residual feature $\bm{X}^{r}$. ${\bm{F}}_{sq}$ is the squeeze function, which is global average pooling here. $S, H, W$ are the number of slices, height, and width of $\bm{X}^{r}$ respectively. ${\bm{F}}_{ex}$ is the excitation function, which is parameterized by two layer fully connected neural networks here with activation functions $\bm{G}$ and $\bm{\sigma}$, and weights ${\bm{W}}_1$ and ${\bm{W}}_2$. The $\bm{\sigma}$ is the sigmoid function. The $\bm{G}$ is typically a ReLU function, but we use LeakyReLU in the AnatomyNet \cite{maas2013rectifier}. We use the learned scale value $s_{k}$ to calibrate the residual feature channel ${\bm{X}^{r}}_{k}$, and obtain the calibrated residual feature $\tilde{\bm{X}}$ . The SE block is illustrated in the upper right corner in Fig. \ref{fig:seunet}.

The AnatomyNet replaces the standard convolutional layers in the U-Net with SE residual blocks to learn effective features. The input of AnatomyNet is a cropped whole-volume head and neck CT image. We remove the down-sampling layers in the second, third, and fourth encoder blocks to improve the performance of segmenting small anatomies. In the output block, we concatenate the input with the transposed convolution feature maps obtained from the second last block. After that, a convolutional layer with 16 $3 \times 3 \times 3$ kernels and LeakyReLU activation function is employed. In the last layer, we use a convolutional layer with 10 $3 \times 3 \times 3$ kernels and soft-max activation function to generate the segmentation probability maps for nine OARs plus background. 

\subsection{Loss function}\label{sec:extremeimbalance}
\begin{figure} [ht]
	\begin{center}
		\begin{tabular}{c} 
				\includegraphics[height=5.7cm]{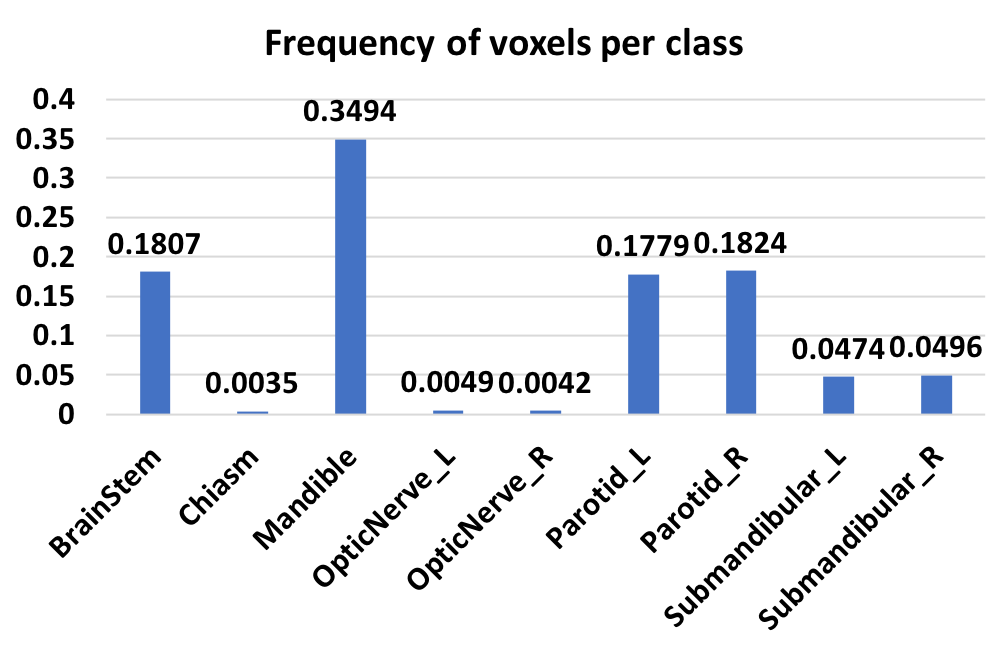}
		\end{tabular}
	\end{center}
	\caption[pixeldist] 
	{ \label{fig:pixeldist} 
		The frequency of voxels for each class on MICCAI 2015 challenge dataset. Background takes up 98.18\% of all the voxels. Chiasm takes only 0.35\% of the foreground which means it only takes about 1/100,000 of the whole-volume CT image. The huge imbalance of voxels in small-volumed organs causes difficulty for small-volumed organ segmentation. }
\end{figure}

Small object segmentation is always a challenge in semantic segmentation. From the learning perspective, the challenge is caused by imbalanced data distribution, because image semantic segmentation requires pixel-wise labeling and small-volumed organs contribute less to the loss. In our case, the small-volumed organs, such as optic chiasm, only take about 1/100,000 of the whole-volume CT images from Fig. \ref{fig:pixeldist}. The dice loss, the minus of dice coefficient (DSC), can be employed to partly address the problem by turning pixel-wise labeling problem into minimizing class-level distribution distance \cite{salehi2017tversky}. 

Several methods have been proposed to alleviate the small-volumed organ segmentation problem. The generalized dice loss uses squared volume weights. However, it makes the optimization unstable in the extremely unbalanced segmentation \cite{sudre2017generalised}. The exponential logarithmic loss is inspired by the focal loss for class-level loss as $\mathbb{E}[{(-\ln (D))}^{\gamma}]$, where $D$ is the dice coefficient (DSC) for the interested class, $\gamma$ can be set as 0.3, and $\mathbb{E}$ is the expectation over classes and whole-volume CT images. The gradient of exponential logarithmic loss w.r.t. DSC $D$ is $-\frac{0.3}{D {\ln (D)}^{0.7}}$. The absolute value of gradient is getting bigger for well-segmented class ($D$ close to 1).  Therefore, the exponential logarithmic loss still places more weights on well-segmented class, and is not effective in learning to improve on not-well-segmented class.

In the AnatomyNet, we employ a hybrid loss consisting of contributions from both dice loss and focal loss \cite{lin2017focal}. The dice loss learns the class distribution alleviating the imbalanced voxel problem, where as the focal loss forces the model to learn poorly classified voxels better. The total loss can be formulated as
\begin{equation}
\label{eq:hybridloss}
\begin{aligned}
{TP}_p(c) =& \sum_{n=1}^{N}{p_{n}(c)g_{n}(c)} \\
{FN}_p(c) =& \sum_{n=1}^{N}{(1-p_{n}(c))g_{n}(c)} \\
{FP}_p(c) =& \sum_{n=1}^{N}{p_{n}(c)(1-g_{n}(c))} \\
\mathcal{L} =& \mathcal{L}_{Dice} + \lambda \mathcal{L}_{Focal} \\
=&C - \sum_{c=0}^{C-1}{\frac{{TP}_p(c)}{{TP}_p(c)
		+ \alpha {FN}_p(c) + \beta {FP}_p(c)}} \\
&- \lambda \frac{1}{N}\sum_{c=0}^{C-1}\sum_{n=1}^{N}g_n(c){(1-p_n(c))}^{2}\log(p_n(c)) \, ,
\end{aligned}
\end{equation}
where ${TP}_p(c)$, ${FN}_p(c)$ and ${FP}_p(c)$ are the true positives, false negatives and false positives for class $c$ calculated by prediction probabilities respectively, $p_n(c)$ is the predicted probability for voxel $n$ being class $c$, $g_n(c)$ is the ground truth for voxel $n$ being class $c$, $C$ is the total number of anatomies plus one (background), $\lambda$ is the trade-off between dice loss $\mathcal{L}_{Dice}$ and focal loss $\mathcal{L}_{Focal}$, $\alpha$ and $\beta$ are the trade-offs of penalties for false negatives and false positives which are set as 0.5 here, $N$ is the total number of voxels in the CT images. $\lambda$ is set to be 0.1, 0.5 or 1 based on the performance on the validation set. Because of size differences for different HaN whole-volume CT images, we set the batch size to be 1. 

\subsection{Handling missing annotations}\label{sec:missingannot}
\vspace{-0.1in}
Another challenge in anatomical segmentation is due to missing annotations common in the training datasets, because annotators often include different anatomies in their annotations. For example, we collect 261 head and neck CT images with anatomical segmentation ground truths from 5 hospitals, and the numbers of nine annotated anatomies are very different as shown in Table \ref{tab:numberanatomy}. To handle this challenge, we mask out the background (denoted as class 0) and the missed anatomy. Let $c \in \{1,2,3,4,5,6,7,8,9\}$ denote the index of anatomies. We employ a mask vector $\bm{m}_i$ for the $i$th CT image, and denote background as label $0$. That is 
$\bm{m}_i(c) = 1$ if anatomy c is annotated, and $0$ otherwise. For the background, the mask is
$\bm{m}_i(0) = 1$ if all anatomies are annotated, and 0 otherwise.
\begin{table}[ht]
	\caption{The numbers of the nine annotated anatomies from 261 head and neck training CT images.} 
	\label{tab:numberanatomy}
	\vspace{-0.1in}
	\begin{center}       
		\begin{tabular}{|c|c|} 
			\hline
			\rule[-1ex]{0pt}{3.5ex}  Anatomy names &\#Annotations\\
			\hline
			\rule[-1ex]{0pt}{3.5ex}  Brain Stem &196\\
			\hline
			\rule[-1ex]{0pt}{3.5ex} Chiasm &129 \\
			\hline
			\rule[-1ex]{0pt}{3.5ex} Mandible &227\\
			\hline
			\rule[-1ex]{0pt}{3.5ex} Opt Ner L &133 \\
			\hline
			\rule[-1ex]{0pt}{3.5ex} Opt Ner R &133 \\
			\hline
			\rule[-1ex]{0pt}{3.5ex} Parotid L &257 \\
			\hline
			\rule[-1ex]{0pt}{3.5ex} Parotid R &256\\
			\hline
			\rule[-1ex]{0pt}{3.5ex} Submand. L &135\\
			\hline
			\rule[-1ex]{0pt}{3.5ex} Submand. R &130 \\
			\hline
		\end{tabular}
	\end{center}
\end{table}
The missing annotations for some anatomical structures cause imbalanced class-level annotations. To address this problem, we employ weighted loss function for balanced weights updating of different anatomies. The weights $\bm{w}$ are set as the inverse of the number of annotations for class $c$, $\bm{w}(c)=1/\sum_{i}\bm{m}_i(c)$, so that the weights in deep networks are updated equally with different anatomies. The dice loss for $i$th CT image in equation \ref{eq:hybridloss} can be written as 
\begin{equation}
\label{eq:missing}
\mathcal{\tilde{L}}_{Dice} =C - \sum_{c=0}^{C-1} \bm{m}_i(c) \bm{w}(c) {\frac{{TP}_p(c)}{{TP}_p(c)
		+ \alpha {FN}_p(c) + \beta {FP}_p(c)}}  \, .
\end{equation}
The focal loss for missing annotations in the $i$th CT image can be written as 
\begin{equation}
\label{eq:focalmissing}
\mathcal{\tilde{L}}_{Focal}
=- \frac{1}{N}\sum_{c=0}^{C-1}\bm{m}_i(c) \bm{w}(c)\sum_{n=1}^{N}g_n(c){(1-p_n(c))}^{2}\log(p_n(c)) \, .
\end{equation}
We use loss $\mathcal{\tilde{L}}_{Dice} + \lambda \mathcal{\tilde{L}}_{Focal}$ in the AnatomyNet.

\subsection{Implementation details and performance evaluation}\label{method_eval}
\vspace{-0.1in}
We implemented AnatomyNet in PyTorch, and trained it on NVIDIA Tesla P40. Batch size was set to be 1 because of different sizes of whole-volume CT images. We first used RMSprop optimizer \cite{tieleman2012lecture} with learning rate being $0.002$ and the number of epochs being 150. Then we used stochastic gradient descend with momentum 0.9, learning rate $0.001$ and the number of epochs 50. During training, we used affine transformation and elastic deformation for data augmentation, implemented on the fly. 

We use Dice coefficient (DSC) as the final evaluation metric, defined to be  $2TP/(2TP+FN+FP)$, where $TP$, $FN$, and $FP$ are true positives, false negatives, false positives, respectively. 

\section{RESULTS}\label{sec:result}
\vspace{-0.1in}
We trained our deep learning model, AnatomyNet, on 261 training samples, and evaluated its performance on the MICCAI head and neck segmentation challenge 2015 test data (10 samples, DATASET 4) and compared it to the performances of previous methods benchmarked on the same test dataset.  Before we present the final results, we first describe the rationale behind several designing choices under AnatomyNet, including architectural designs and model training. 
\vspace{-0.2in}
\subsection{Determining down-sampling scheme}
\vspace{-0.1in}
The standard U-Net model has multiple down-sampling layers, which help the model learn high-level image features.  However, down-sampling also reduces image resolution and makes it harder to segment small OARs such as optic nerves and chiasm.  To evaluate the effect of the number of down-sampling layers on the segmentation performance, we experimented with four different down-sampling schemes shown in Table \ref{tab:netstructpool}.  Pool 1 uses only one down-sampling step, while Pool 2, 3, and 4 use 2, 3 and 4 down-sampling steps, respectively, distributed over consecutive blocks. With each down-sampling, the feature map size is reduced by half.  We incorporated each of the four down-sampling schemes into the standard U-Net model, which was then trained on the training set and evaluated on the test set.  For fair comparisons, we used the same number of filters in each layer. The decoder layers of each model are set to be symmetric with the encoder layers.

\begin{table}[ht]
	\caption{Sizes of encoder blocks in U-Nets with different numbers of down-samplings. The number before symbol @ denotes the number of output channels, while the number after the symbol denotes the size of feature map relative to the input.} 
	\label{tab:netstructpool}
	\vspace{-0.1in}
	\begin{center}       
		\begin{tabular}{|c|c|c|c|c|} 
			\hline
			\rule[-1ex]{0pt}{3.5ex}  Nets &1st block & 2nd block & 3rd block & 4th block\\
			\hline
			\rule[-1ex]{0pt}{3.5ex}  Pool 1 &$32@{(1/2)}^3$&$40@{(1/2)}^3$&$48@{(1/2)}^3$&$56@{(1/2)}^3$\\
			\hline
			\rule[-1ex]{0pt}{3.5ex} Pool 2 &$32@{(1/2)}^3$&$40@{(1/4)}^3$&$48@{(1/4)}^3$&$56@{(1/4)}^3$\\
			\hline
			\rule[-1ex]{0pt}{3.5ex} Pool 3 &$32@{(1/2)}^3$&$40@{(1/4)}^3$&$48@{(1/8)}^3$&$56@{(1/8)}^3$\\
			\hline
			\rule[-1ex]{0pt}{3.5ex} Pool 4 &$32@{(1/2)}^3$&$40@{(1/4)}^3$&$48@{(1/8)}^3$&$56@{(1/16)}^3$\\
			\hline
		\end{tabular}
	\end{center}
\end{table}

The DSC scores of the four down-sampling schemes are shown in Table \ref{tab:numpool}. On average, one down-sampling block (Pool 1) yields the best average performance, beating other down-sampling schemes in 6 out of 9 anatomies. The performance gaps are most prominent on three small-volumed OARs -  optic nerve left, optic nerve right and optic chiasm, which demonstrates that the U-Net with one down-sampling layer works better on small organ segmentation than the standard U-Net. The probable reason is that small organs reside in only a few slices and more down-sampling layers are more likely to miss features for the small organs in the deeper layers. Based on these results, we decide to use only one down-sampling layer in AnatomyNet  (Fig. \ref{fig:seunet}). 

	\begin{table}[ht]
		\caption{Performances of U-Net models with different numbers of down-sampling layers, measured with Dice coefficients.} 
		\label{tab:numpool}
		\vspace{-0.1in}
		\begin{center}    
				\begin{tabular}{|c|c|c|c|c|} 
					\hline
					\rule[-1ex]{0pt}{3.5ex}  Anatomy &Pool 1 &Pool 2 & Pool 3 & Pool 4 \\
					\hline
					\rule[-1ex]{0pt}{3.5ex}  Brain Stem&85.1&\textbf{85.3}&84.3&84.9\\
					\hline
					\rule[-1ex]{0pt}{3.5ex} Chiasm &\textbf{50.1}&48.7&47.0&45.3\\
					\hline
					\rule[-1ex]{0pt}{3.5ex} Mand. &\textbf{91.5}&89.9&90.6&90.1\\
					\hline
					\rule[-1ex]{0pt}{3.5ex} Optic Ner L &\textbf{69.1}&65.7&67.2&67.9\\
					\hline
					\rule[-1ex]{0pt}{3.5ex} Optic Ner R &\textbf{66.9}&65.0&66.2&63.7\\
					\hline
					\rule[-1ex]{0pt}{3.5ex} Paro. L &\textbf{86.6}&\textbf{86.6}&85.9&\textbf{86.6}\\
					\hline
					\rule[-1ex]{0pt}{3.5ex} Paro. R &85.6&85.3&84.8&\textbf{85.9}\\
					\hline
					\rule[-1ex]{0pt}{3.5ex} Subm. L &\textbf{78.5}&77.9&77.9&77.3\\
					\hline
					\rule[-1ex]{0pt}{3.5ex} Subm. R &77.7&\textbf{78.4}&76.6&77.8\\
					\hline
					\rule[-1ex]{0pt}{3.5ex} Average &\textbf{76.8}&75.9&75.6&75.5\\
					\hline
				\end{tabular}
		\end{center}
\end{table}

\vspace{-0.2in}
\subsection{Choosing network structures}\label{sec:expnetstruct}
\vspace{-0.1in}
In addition to down-sample schemes, we also tested several other architecture designing choices. The first one is on how to combine features from horizontal layers within U-Net. Traditional U-Net uses concatenation to combine features from horizontal layers in the decoder, as illustrated with dash lines in Fig. \ref{fig:seunet}. However, recent feature pyramid network (FPN) recommends summation to combine horizontal features \cite{lin2017feature}.   Another designing choice is on choosing local feature learning blocks with each layer.  The traditional U-Net uses simple 2D convolution, extended to 3D convolution in our case.  To learn more effective features, we tried two other feature learning blocks: a) residual learning, and b) squeeze-and-excitation residual learning.  Altogether, we investigated the performances of the following six architectural designing choices:
\begin{enumerate}
\vspace{-0.1in}
\item \textbf{3D SE Res UNet}, the architecture implemented in AnatomyNet (Fig.\ \ref{fig:seunet}) with both squeeze-excitation residual learning and concatenated horizontal features. 
\vspace{-0.1in}
\item \textbf{3D Res UNet}, replacing the SE Residual blocks in 3D SE Res UNet with residual blocks.
\vspace{-0.1in}
\item \textbf{Vanilla U-Net}, replacing the SE Residual blocks in 3D SE Res UNet with 3D convolutional layers.
\vspace{-0.1in}
\item \textbf{3D SE Res UNet (sum)}, replacing concatenations in 3D SE Res UNet with summations. When the numbers of channels are different, one additional $1\times 1 \times 1$ 3D convolutional layer is used to map the encoder to the same size as the decoder.
\vspace{-0.1in}
\item \textbf{3D Res UNet (sum)}, replacing the SE Residual blocks in 3D SE Res UNet (sum) with residual blocks.
\vspace{-0.1in}
\item \textbf{Vanilla U-Net (sum)}, replacing the SE Residual blocks in 3D SE Res UNet (sum) with 3D convolutional layers.
\end{enumerate}

The six models were trained on the same training dataset with identical training procedures. The performances measured by DSC on the  test dataset are summarized in Table \ref{tab:seres_sum}. We notice a few observations from this study. First, feature concatenation shows consistently better performance than feature summation. It seems feature concatenation provides more flexibility in feature learning than the fixed operation through feature summation. Second, 3D SE residual U-Net with concatenation yields the best performance. It demonstrates the power of SE features on 3D semantic segmentation, because the SE scheme learns the channel-wise calibration and helps alleviate the dependencies among channel-wise features as discussed in Section \ref{sec:anatomynet}. 

The SE residual block learning incorporated in AnatomyNet results in 2-3\%  improvements in DSC over the traditional U-Net model, outperforming U-Net in 6 out of 9 anatomies. 
\vspace{-0.1in}
	\begin{table}[ht]
		\caption{Performance comparison on different network structures} 
		\label{tab:seres_sum}
		\vspace{-0.1in}
		\begin{center}    
			\resizebox{\columnwidth}{!}{   
				\begin{tabular}{|c|c|c|c|c|c|c|} 
					\hline
					\rule[-1ex]{0pt}{3.5ex}  Anatomy &\begin{tabular}{@{}c@{}}Vanilla \\ UNet\end{tabular} & \begin{tabular}{@{}c@{}}3D Res\\ UNet\end{tabular} & \begin{tabular}{@{}c@{}}3D SE Res \\ UNet\end{tabular} & \begin{tabular}{@{}c@{}}Vanilla \\ UNet (sum)\end{tabular} & \begin{tabular}{@{}c@{}}3D Res \\ UNet (sum)\end{tabular}&\begin{tabular}{@{}c@{}}3D SE Res \\ UNet (sum)\end{tabular}\\
					\hline
					\rule[-1ex]{0pt}{3.5ex}  \begin{tabular}{@{}c@{}}Brain \\ Stem\end{tabular}&85.1&85.9&\textbf{86.4}&85.0&85.8&86.0\\
					\hline
					\rule[-1ex]{0pt}{3.5ex} Chiasm &50.1&53.3&53.2&50.8&49.8&\textbf{53.5}\\
					\hline
					\rule[-1ex]{0pt}{3.5ex} Mand. &91.5&90.6&\textbf{92.3}&90.2&90.9&91.3\\
					\hline
					\rule[-1ex]{0pt}{3.5ex} \begin{tabular}{@{}c@{}}Optic \\ Ner L\end{tabular} &69.1&69.8&\textbf{72.0}&66.5&70.4&68.9\\
					\hline
					\rule[-1ex]{0pt}{3.5ex} \begin{tabular}{@{}c@{}}Optic \\ Ner R\end{tabular} &66.9&67.5&\textbf{69.1}&66.1&67.4&45.6\\
					\hline
					\rule[-1ex]{0pt}{3.5ex} Paro. L &86.6&\textbf{87.8}&\textbf{87.8}&86.8&87.5&87.4\\
					\hline
					\rule[-1ex]{0pt}{3.5ex} Paro. R &85.6&86.2&86.8&86.0&86.7&\textbf{86.9}\\
					\hline
					\rule[-1ex]{0pt}{3.5ex} Subm. L &78.5&79.9&81.1&79.3&79.1&\textbf{82.4}\\
					\hline
					\rule[-1ex]{0pt}{3.5ex} Subm. R &77.7&80.2&\textbf{80.8}&77.6&78.4&80.5\\
					\hline
					\rule[-1ex]{0pt}{3.5ex} Average &76.8&77.9&\textbf{78.8}&76.5&77.3&75.8\\
					\hline
				\end{tabular}
			}
		\end{center}
\end{table}

\vspace{-0.2in}
\subsection{Choosing loss functions}
We also validated the effects of different loss functions on training and model performance. To differentiate the effects of loss functions from network design choices,  we used only the vanilla U-Net and trained it with different loss functions. This way, we can focus on studying the impact of loss functions on model performances.  We tried four loss functions, including Dice loss, exponential logarithmic loss, hybrid loss between Dice loss and focal loss, and hybrid loss between Dice loss and cross entropy. The trade-off parameter in hybrid losses ($\lambda$ in Eq.\ \ref{eq:hybridloss}) was chosen from either 0.1, 0.5 or 1, based on the performance on a validation set. For hybrid loss between Dice loss and focal loss, the best $\lambda$ was found to be 0.5. For hybrid loss between Dice loss and cross entropy, the best $\lambda$ was 0.1.


\vspace{-0.2in}
	\begin{table}[ht]
		\caption{Comparisons of test performances of models trained with different loss functions, evaluated with Dice coefficients.} 
		\label{tab:lossfunc}
		\vspace{-0.1in}
		\begin{center}    
				\begin{tabular}{|c|c|c|c|c|} 
					\hline
					\rule[-1ex]{0pt}{3.5ex}  Anatomy &\begin{tabular}{@{}c@{}}Dice \\ loss\end{tabular} & \begin{tabular}{@{}c@{}}Exp. Log.\\ Dice \end{tabular} & \begin{tabular}{@{}c@{}}Dice + \\ focal \end{tabular} & \begin{tabular}{@{}c@{}}Dice + cross \\ entropy \end{tabular}\\
					\hline
					\rule[-1ex]{0pt}{3.5ex}  Brain Stem&85.1&85.0&\textbf{86.1}&85.2\\
					\hline
					\rule[-1ex]{0pt}{3.5ex} Chiasm &50.1&50.0&\textbf{52.2}&48.8\\
					\hline
					\rule[-1ex]{0pt}{3.5ex} Mand. &\textbf{91.5}&89.9&90.0&91.0\\
					\hline
					\rule[-1ex]{0pt}{3.5ex} Optic Ner L &69.1&67.9&68.4&\textbf{69.6}\\
					\hline
					\rule[-1ex]{0pt}{3.5ex} Optic Ner R &66.9&65.9&\textbf{69.1}&67.4\\
					\hline
					\rule[-1ex]{0pt}{3.5ex} Paro. L &86.6&86.4&87.4&\textbf{88.0}\\
					\hline
					\rule[-1ex]{0pt}{3.5ex} Paro. R &85.6&84.8&86.3&\textbf{86.9}\\
					\hline
					\rule[-1ex]{0pt}{3.5ex} Subm. L &78.5&76.3&\textbf{79.6}&77.8\\
					\hline
					\rule[-1ex]{0pt}{3.5ex} Subm. R &77.7&78.2&\textbf{79.8}&78.4\\
					\hline
					\rule[-1ex]{0pt}{3.5ex} Average &76.8&76.0&\textbf{77.7}&77.0\\
					\hline
				\end{tabular}
		\end{center}
\end{table}

The performances of the model trained with the four loss functions described above are shown in Table \ref{tab:lossfunc}. The performances are measured in terms of the average DSC on the test dataset. We notice a few observations from this experiment. First, the two hybrid loss functions consistently outperform simple Dice or exponential logarithmic loss, beating the other two losses in 8 out of 9 anatomies. This suggests that taking the voxel-level loss into account can improve performance. Second, between the two hybrid losses, Dice combined with focal loss has better performances.  In particular, it leads to significant improvements (2-3\%) on segmenting two small anatomies - optic nerve R and optic chiasm, consistent with our motivation discussed in the Section \ref{sec:extremeimbalance}.  

Based on the above observations, the hybrid loss with Dice combined with focal loss was used to train AnatomyNet, and benchmark its performance against previous methods.

\vspace{-0.2in}
\subsection{Comparing to state-of-the-art methods}
After having determined the structure of AnatomyNet and the loss function for training it, we set out to compare its performance with previous state-of-the-art methods. For consistency purpose, all models were evaluated on the MICCAI head and neck challenge 2015 test set. The average DSC of different methods are summarized in Table \ref{tab:anatomynet}. The best result for each anatomy from the MICCAI 2015 challenge is denoted as MICCAI 2015 \cite{raudaschl2017evaluation}, which may come from different teams with different methods. 
\vspace{-0.2in}
\begin{table}[ht]
	\caption{Performance comparisons with state-of-the-art methods, showing the average DSC on the test set.} 
	\label{tab:anatomynet}
	\vspace{-0.2in}
	\begin{center}    
		\resizebox{\columnwidth}{!}{   
		\begin{tabular}{|c|c|c|c|c|c|} 
			\hline
			\rule[-1ex]{0pt}{3.5ex}  Anatomy &\begin{tabular}{@{}c@{}}\textbf{MICCAI} \\ \textbf{2015} \cite{raudaschl2017evaluation}\end{tabular} & \begin{tabular}{@{}c@{}}Fritscher et al. \\ 2016 \cite{fritscher2016deep}\end{tabular} & \begin{tabular}{@{}c@{}}Ren et al. \\ 2018 \cite{ren2018interleaved}\end{tabular} & \begin{tabular}{@{}c@{}}Wang et al. \\ 2018 \cite{wang2018hierarchical}\end{tabular} & AnatomyNet\\
			\hline
			\rule[-1ex]{0pt}{3.5ex}  \begin{tabular}{@{}c@{}}Brain \\ Stem\end{tabular}&88&N/A&N/A&\textbf{90$\pm$4}&86.65$\pm$2\\
			\hline
			\rule[-1ex]{0pt}{3.5ex} Chiasm &55&49$\pm$9&\textbf{58$\pm$17}&N/A&53.22$\pm$15\\
			\hline
			\rule[-1ex]{0pt}{3.5ex} Mand. &93&N/A&N/A&\textbf{94}$\pm$1&92.51$\pm$2\\
			\hline
			\rule[-1ex]{0pt}{3.5ex} \begin{tabular}{@{}c@{}}Optic \\ Ner L\end{tabular} &62&N/A&72$\pm$8&N/A&\textbf{72.10$\pm$6}\\
			\hline
			\rule[-1ex]{0pt}{3.5ex} \begin{tabular}{@{}c@{}}Optic \\ Ner R\end{tabular} &62&N/A&70$\pm$9&N/A&\textbf{70.64$\pm$10}\\
			\hline
			\rule[-1ex]{0pt}{3.5ex} Paro. L &84&81$\pm$4&N/A&83$\pm$6&\textbf{88.07$\pm$2}\\
			\hline
			\rule[-1ex]{0pt}{3.5ex} Paro. R &84&81$\pm$4&N/A&83$\pm$6&\textbf{87.35$\pm$4}\\
			\hline
			\rule[-1ex]{0pt}{3.5ex} Subm. L &78&65$\pm$8&N/A&N/A&\textbf{81.37$\pm$4}\\
			\hline
			\rule[-1ex]{0pt}{3.5ex} Subm. R &78&65$\pm$8&N/A&N/A&\textbf{81.30$\pm$4}\\
			\hline
			\rule[-1ex]{0pt}{3.5ex} Average &76&N/A&N/A&N/A&\textbf{79.25}\\
			\hline
		\end{tabular}
	}
	\end{center}
\end{table}

MICCAI 2015 competition merged left and right paired organs into one target, while we treat them as two separate anatomies.  As a result, MICCAI 2015 competition is a seven (6 organs + background) class segmentation and ours is a ten-class segmentation, which makes the segmentation task more challenging. Nonetheless, the AnatomyNet achieves an average Dice coefficient of 79.25, which is 3.3\% better than the best result from MICCAI 2015 Challenge (Table \ref{tab:anatomynet}). In particular, the improvements on optic nerves are about 9-10\%, suggesting that deep learning models are better equipped to handle small anatomies with large variations among patients.   The AnatomyNet also outperforms the atlas based ConvNets in \cite{fritscher2016deep} on all classes, which is likely contributed by the fact that the end-to-end structure in AnatomyNet for whole-volume HaN CT image captures global information for relative spatial locations among anatomies. Compared to the interleaved ConvNets in \cite{ren2018interleaved} on small-volumed organs, such as chiasm, optic nerve left and optic nerve right, AnatomyNet is better on 2 out of 3 cases. The interleaved ConvNets achieved higher performance on chiasm, which is likely contributed by the fact that its prediction was operated on small region of interest (ROI), obtained first through atlas registration, while AnatomyNet operates directly on whole-volume slices. 

Aside from the improvement on segmentation accuracy, another advantage of AnatomyNet is that it is orders of magnitude faster than traditional atlas-based methods using in the MICCAI 2015 challenge.  AnatomyNet takes about 0.12 seconds to fully segment  a head and neck CT image of  dimension  $178 \times 302 \times 225$. By contrast, the atlas-based methods can take a dozen minutes to complete one segmentation depending on implementation details and the choices on the number of atlases. 


\vspace{-0.2in}
\subsection{Visualizations on MICCAI 2015 test}
\begin{figure} [ht]
	\begin{center}
		 		\begin{minipage}{0.23\linewidth}
		 			\includegraphics[width=\textwidth]{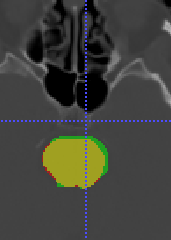}
		 		\end{minipage}
	 		\begin{minipage}{0.23\linewidth}
	 			\includegraphics[width=\textwidth]{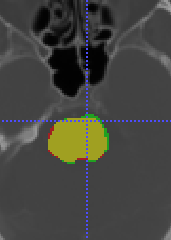}
	 		\end{minipage}
 		\begin{minipage}{0.23\linewidth}
 			\includegraphics[width=\textwidth]{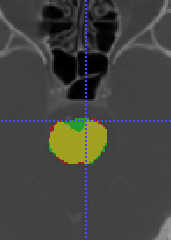}
 		\end{minipage}
 		\begin{minipage}{0.23\linewidth}
 			\includegraphics[width=\textwidth]{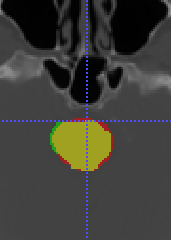}
 		\end{minipage} \\
 		
		 		\begin{minipage}{0.23\linewidth}
		 			\includegraphics[width=\textwidth]{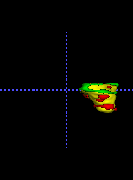}
		 		\end{minipage}
	 		\begin{minipage}{0.23\linewidth}
	 			\includegraphics[width=\textwidth]{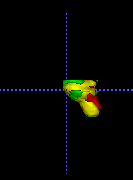}
	 		\end{minipage}
 		\begin{minipage}{0.23\linewidth}
 			\includegraphics[width=\textwidth]{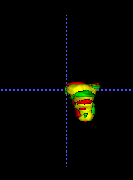}
 		\end{minipage}
 	\begin{minipage}{0.23\linewidth}
 		\includegraphics[width=\textwidth]{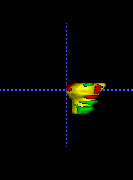}
 	\end{minipage} \\
 	
		 		\begin{minipage}{0.23\linewidth}
		 			\includegraphics[width=\textwidth]{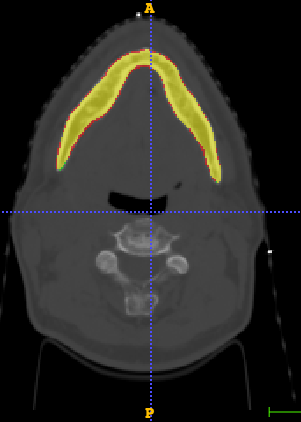}
		 		\end{minipage}
	 		\begin{minipage}{0.23\linewidth}
	 			\includegraphics[width=\textwidth]{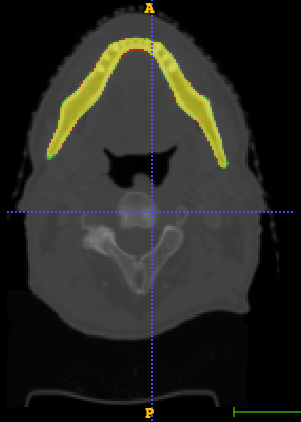}
	 		\end{minipage}
 		\begin{minipage}{0.23\linewidth}
 			\includegraphics[width=\textwidth]{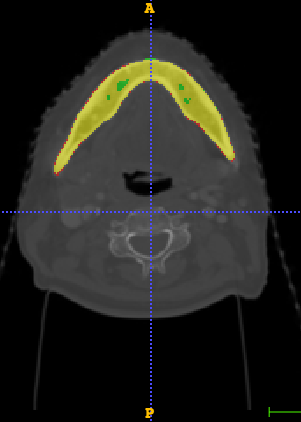}
 		\end{minipage}
 	\begin{minipage}{0.23\linewidth}
 		\includegraphics[width=\textwidth]{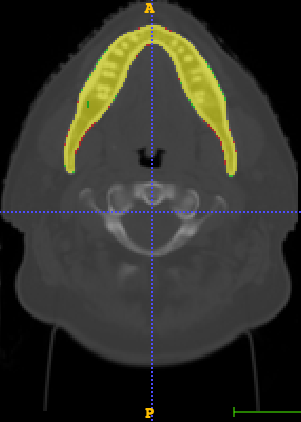}
 	\end{minipage}\\
 
		 		\begin{minipage}{0.23\linewidth}
		 			\includegraphics[width=\textwidth]{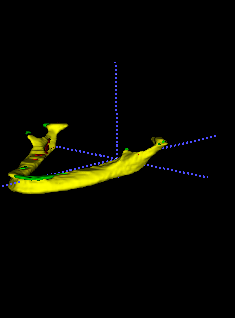}
		 		\end{minipage}
	 		\begin{minipage}{0.23\linewidth}
	 			\includegraphics[width=\textwidth]{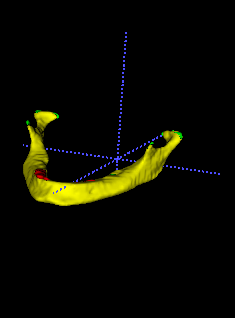}
	 		\end{minipage}
 		\begin{minipage}{0.23\linewidth}
 			\includegraphics[width=\textwidth]{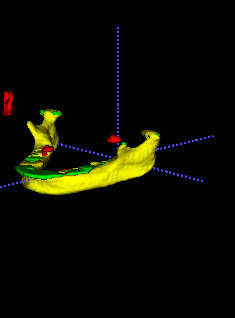}
 		\end{minipage}
 	\begin{minipage}{0.23\linewidth}
 		\includegraphics[width=\textwidth]{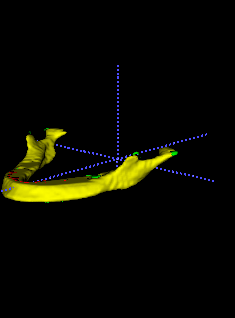}
 	\end{minipage} \\
 
		 		\begin{minipage}{0.23\linewidth}
		 			\includegraphics[width=\textwidth]{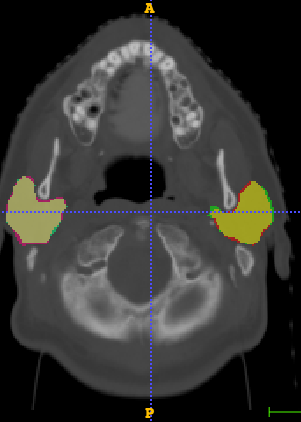}
		 		\end{minipage}
	 		\begin{minipage}{0.23\linewidth}
	 			\includegraphics[width=\textwidth]{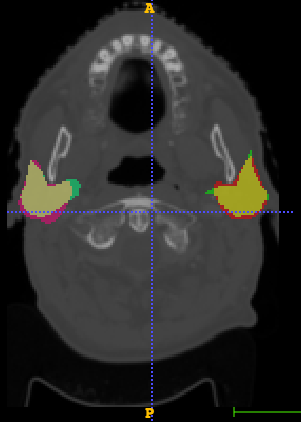}
	 		\end{minipage}
 		\begin{minipage}{0.23\linewidth}
 			\includegraphics[width=\textwidth]{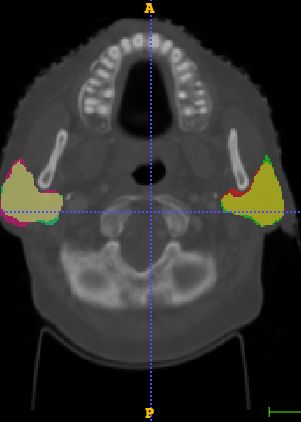}
 		\end{minipage}
 	\begin{minipage}{0.23\linewidth}
 		\includegraphics[width=\textwidth]{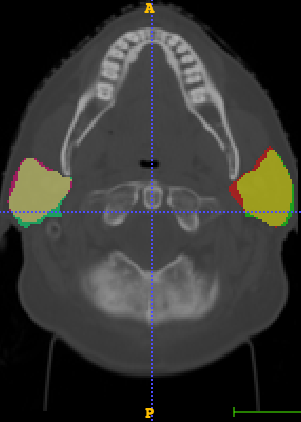}
 	\end{minipage} \\
 
		 		\begin{minipage}{0.23\linewidth}
		 			\includegraphics[width=\textwidth]{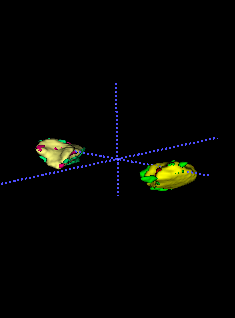}
		 		\end{minipage}
	 		\begin{minipage}{0.23\linewidth}
	 			\includegraphics[width=\textwidth]{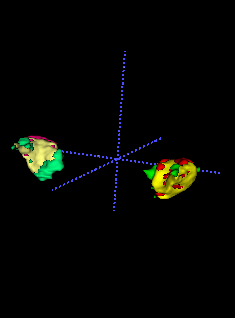}
	 		\end{minipage}
 		\begin{minipage}{0.23\linewidth}
 			\includegraphics[width=\textwidth]{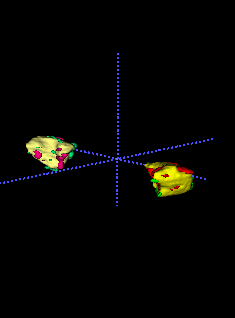}
 		\end{minipage}
 	\begin{minipage}{0.23\linewidth}
 		\includegraphics[width=\textwidth]{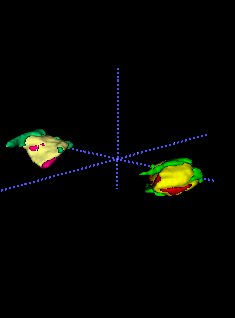}
 	\end{minipage} 
	\end{center}
\vspace{-0.2in}
	\caption[vis1] 
	{ \label{fig:vis1} 
	Visualizations on four test CT images. Rows from top to bottom represent brain stem, brain stem 3D, mandibular, mandibular 3D, parotid left and right, and parotid left and right 3D. Each column represents one CT image. Green represents ground truths, and red denotes predictions. Yellow is the overlap. } 
\end{figure}

\begin{figure} [ht]
	\begin{center}
				 		\begin{minipage}{0.23\linewidth}
			\includegraphics[width=\textwidth]{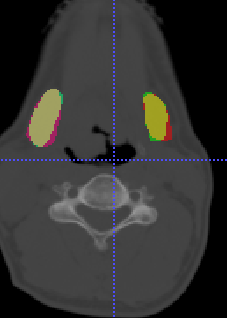}
		\end{minipage}
		\begin{minipage}{0.23\linewidth}
			\includegraphics[width=\textwidth]{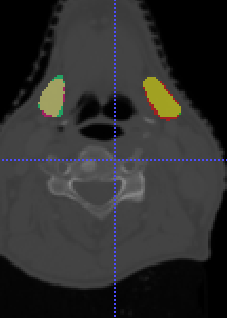}
		\end{minipage}
		\begin{minipage}{0.23\linewidth}
			\includegraphics[width=\textwidth]{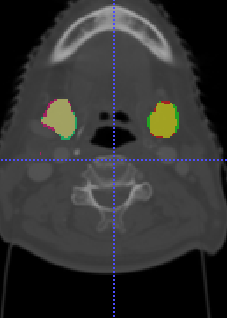}
		\end{minipage}
		\begin{minipage}{0.23\linewidth}
			\includegraphics[width=\textwidth]{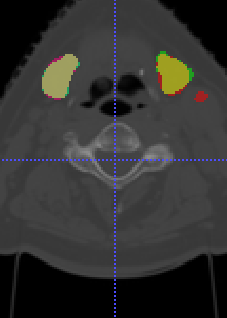}
		\end{minipage}\\
		
		\begin{minipage}{0.23\linewidth}
			\includegraphics[width=\textwidth]{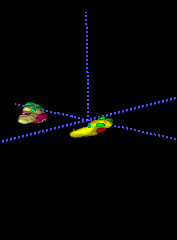}
		\end{minipage}
		\begin{minipage}{0.23\linewidth}
			\includegraphics[width=\textwidth]{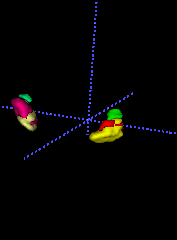}
		\end{minipage}
		\begin{minipage}{0.23\linewidth}
			\includegraphics[width=\textwidth]{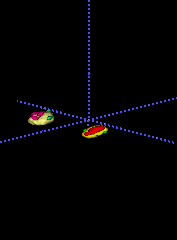}
		\end{minipage}
		\begin{minipage}{0.23\linewidth}
			\includegraphics[width=\textwidth]{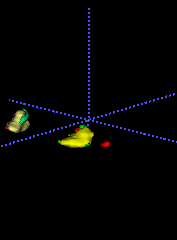}
		\end{minipage}  \\
	
		\begin{minipage}{0.23\linewidth}
			\includegraphics[width=\textwidth]{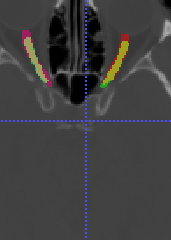}
		\end{minipage}
		\begin{minipage}{0.23\linewidth}
			\includegraphics[width=\textwidth]{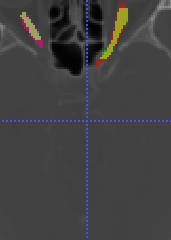}
		\end{minipage}
		\begin{minipage}{0.23\linewidth}
			\includegraphics[width=\textwidth]{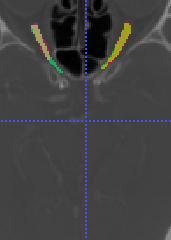}
		\end{minipage}
		\begin{minipage}{0.23\linewidth}
			\includegraphics[width=\textwidth]{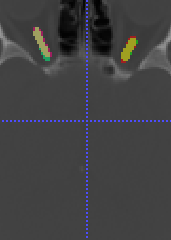}
		\end{minipage}\\
		
		\begin{minipage}{0.23\linewidth}
			\includegraphics[width=\textwidth]{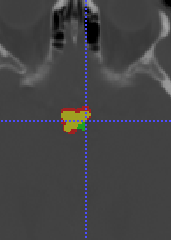}
		\end{minipage}
		\begin{minipage}{0.23\linewidth}
			\includegraphics[width=\textwidth]{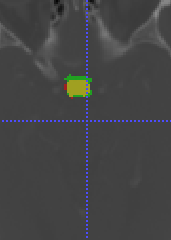}
		\end{minipage}
		\begin{minipage}{0.23\linewidth}
			\includegraphics[width=\textwidth]{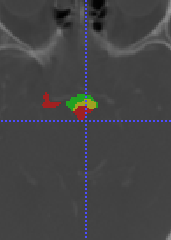}
		\end{minipage}
		\begin{minipage}{0.23\linewidth}
			\includegraphics[width=\textwidth]{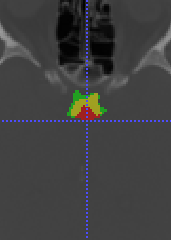}
		\end{minipage}
	\end{center}
\vspace{-0.2in}
	\caption[vis1_2] 
	{ \label{fig:vis1_2} 
		Visualizations on four test CT images. Rows from top to bottom represent submandibular left and right, submandibular left and right 3D, optic nerve left and right, and chiasm. Each column represents one CT image. Green represents ground truths, and red denotes predictions. Yellow is the overlap. The AnatomyNet performs well on small-volumed anatomies.} 
\end{figure}

\vspace{-0.1in}
In Fig.\ \ref{fig:vis1} and Fig.\ \ref{fig:vis1_2}, we visualize the segmentation results by AnatomyNet on four cases from the test datset.   Each row represents one (left and right) anatomy or 3D reconstructed anatomy. Each column denotes one sample. The last two columns show cases where AnatomyNet did not perform well.  The discussions of these cases are presented in Section \ref{sec:futurework}. Green denotes the ground truth. Red represents predicted segmentation results. Yellow denotes the overlap between ground truth and prediction. We visualize the slices containing the largest area of each related organ.  For small OARs such as optic nerves and chiasm (shown in  Fig.\ \ref{fig:vis1_2}), only cross-sectional slices are shown. 


\vspace{-0.2in}
\subsection{Visualizations on independent samples}\label{sec:holdout}
\begin{figure} [ht]
	\begin{center}
		 		\begin{minipage}{0.23\linewidth}
		 			\includegraphics[width=\textwidth]{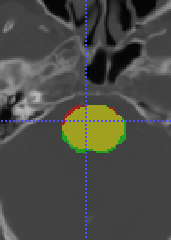}
		 		\end{minipage}
	 		\begin{minipage}{0.23\linewidth}
	 			\includegraphics[width=\textwidth]{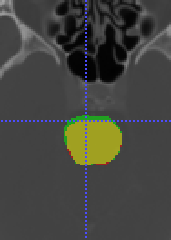}
	 		\end{minipage}
 		\begin{minipage}{0.23\linewidth}
 			\includegraphics[width=\textwidth]{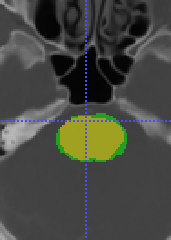}
 		\end{minipage}
 	\begin{minipage}{0.23\linewidth}
 		\includegraphics[width=\textwidth]{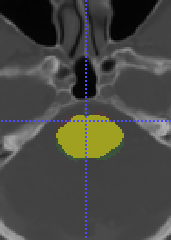}
 	\end{minipage}\\
 
		 		\begin{minipage}{0.23\linewidth}
		 			\includegraphics[width=\textwidth]{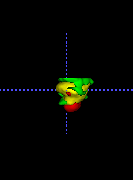}
		 		\end{minipage}
	 		\begin{minipage}{0.23\linewidth}
	 			\includegraphics[width=\textwidth]{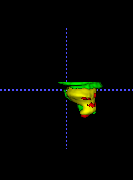}
	 		\end{minipage}
 		\begin{minipage}{0.23\linewidth}
 			\includegraphics[width=\textwidth]{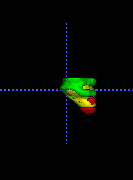}
 		\end{minipage}
 	\begin{minipage}{0.23\linewidth}
 		\includegraphics[width=\textwidth]{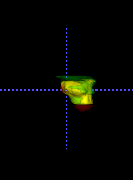}
 	\end{minipage}\\
 
		 		\begin{minipage}{0.23\linewidth}
		 			\includegraphics[width=\textwidth]{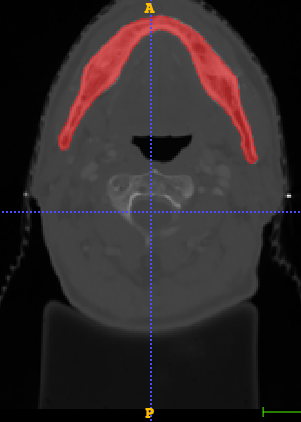}
		 		\end{minipage}
	 		\begin{minipage}{0.23\linewidth}
	 			\includegraphics[width=\textwidth]{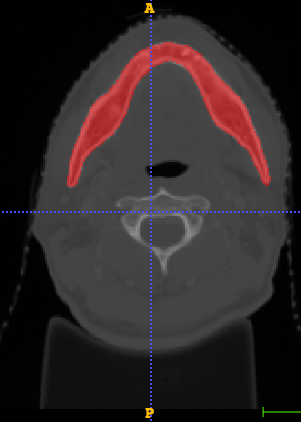}
	 		\end{minipage}
 		\begin{minipage}{0.23\linewidth}
 			\includegraphics[width=\textwidth]{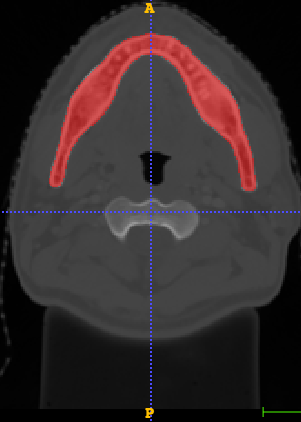}
 		\end{minipage}
 	\begin{minipage}{0.23\linewidth}
 		\includegraphics[width=\textwidth]{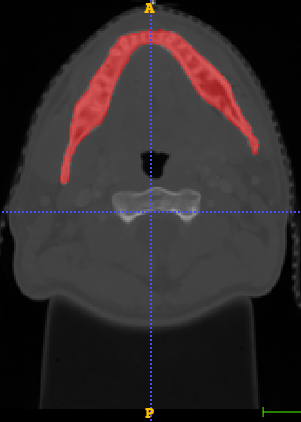}
 	\end{minipage}\\
 
		 		\begin{minipage}{0.23\linewidth}
		 			\includegraphics[width=\textwidth]{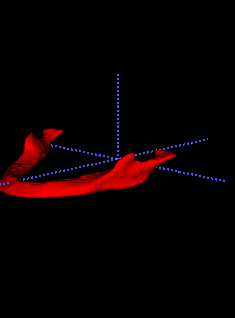}
		 		\end{minipage}
	 		\begin{minipage}{0.23\linewidth}
	 			\includegraphics[width=\textwidth]{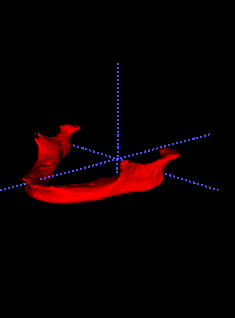}
	 		\end{minipage}
 		\begin{minipage}{0.23\linewidth}
 			\includegraphics[width=\textwidth]{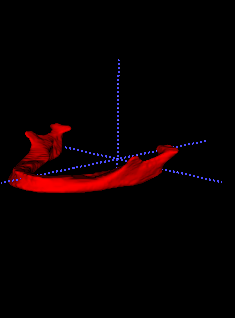}
 		\end{minipage}
 	\begin{minipage}{0.23\linewidth}
 		\includegraphics[width=\textwidth]{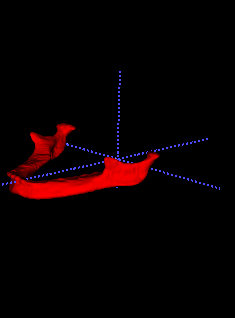}
 	\end{minipage}\\
 
		 		\begin{minipage}{0.23\linewidth}
		 			\includegraphics[width=\textwidth]{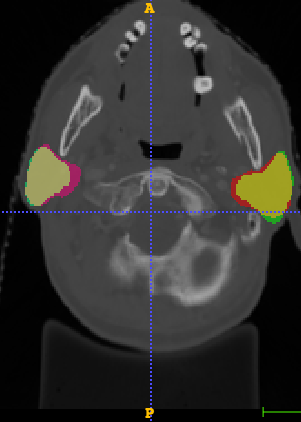}
		 		\end{minipage}
	 		\begin{minipage}{0.23\linewidth}
	 			\includegraphics[width=\textwidth]{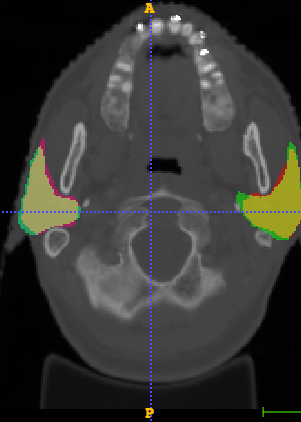}
	 		\end{minipage}
 		\begin{minipage}{0.23\linewidth}
 			\includegraphics[width=\textwidth]{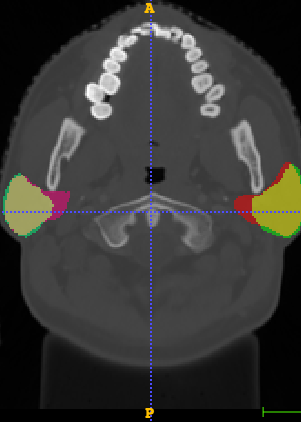}
 		\end{minipage}
 	\begin{minipage}{0.23\linewidth}
 		\includegraphics[width=\textwidth]{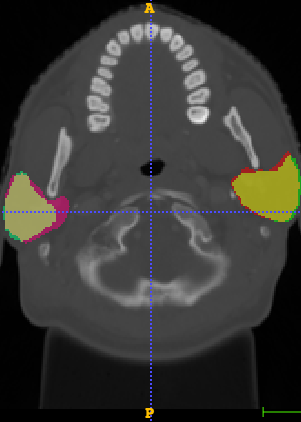}
 	\end{minipage}\\
 
		 		\begin{minipage}{0.23\linewidth}
		 			\includegraphics[width=\textwidth]{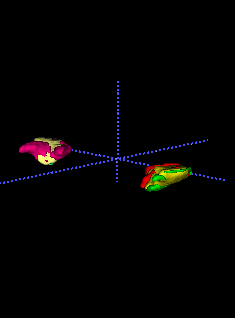}
		 		\end{minipage}
	 		\begin{minipage}{0.23\linewidth}
	 			\includegraphics[width=\textwidth]{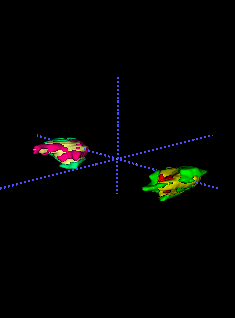}
	 		\end{minipage}
 		\begin{minipage}{0.23\linewidth}
 			\includegraphics[width=\textwidth]{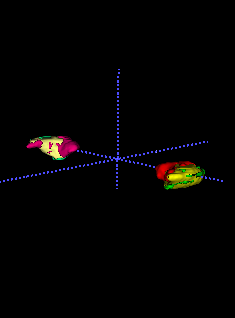}
 		\end{minipage}
 	\begin{minipage}{0.23\linewidth}
 		\includegraphics[width=\textwidth]{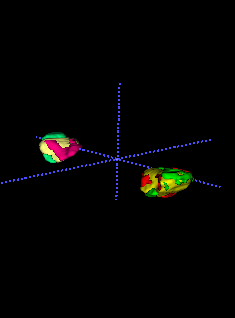}
 	\end{minipage}
	\end{center}
\vspace{-0.2in}
	\caption[vis2] 
	{ \label{fig:vis2} 
	Visualizations for the first four anatomy on the first four holdout CT images. There is no ground truth for mandible and submandibular glands. Because this is a different source from MICCAI 2015, the annotations of brain stem and chiasm are inconsistent with those from MICCAI 2015. The AnatomyNet generalizes well for hold out test set.}
\end{figure}

\begin{figure} [ht]
	\begin{center}
		\begin{minipage}{0.23\linewidth}
			\includegraphics[width=\textwidth]{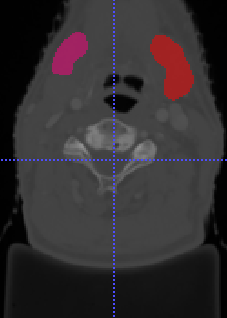}
		\end{minipage}
		\begin{minipage}{0.23\linewidth}
			\includegraphics[width=\textwidth]{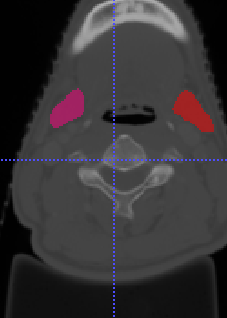}
		\end{minipage}
		\begin{minipage}{0.23\linewidth}
			\includegraphics[width=\textwidth]{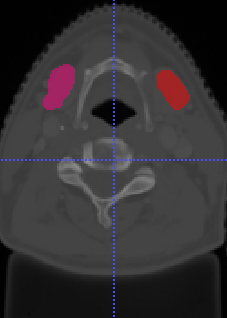}
		\end{minipage}
		\begin{minipage}{0.23\linewidth}
			\includegraphics[width=\textwidth]{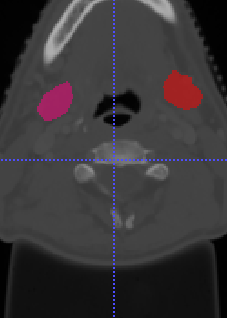}
		\end{minipage}\\
		
		\begin{minipage}{0.23\linewidth}
			\includegraphics[width=\textwidth]{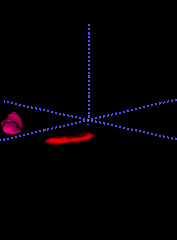}
		\end{minipage}
		\begin{minipage}{0.23\linewidth}
			\includegraphics[width=\textwidth]{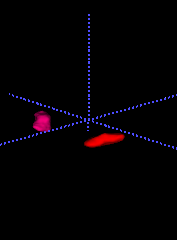}
		\end{minipage}
		\begin{minipage}{0.23\linewidth}
			\includegraphics[width=\textwidth]{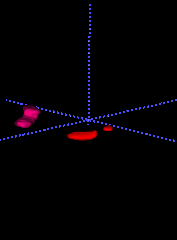}
		\end{minipage}
		\begin{minipage}{0.23\linewidth}
			\includegraphics[width=\textwidth]{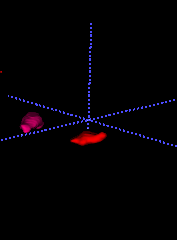}
		\end{minipage}\\
		
		\begin{minipage}{0.23\linewidth}
			\includegraphics[width=\textwidth]{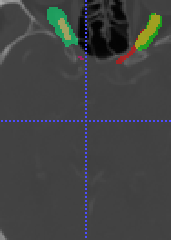}
		\end{minipage}
		\begin{minipage}{0.23\linewidth}
			\includegraphics[width=\textwidth]{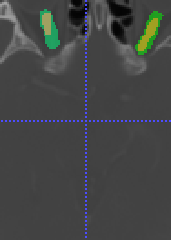}
		\end{minipage}
		\begin{minipage}{0.23\linewidth}
			\includegraphics[width=\textwidth]{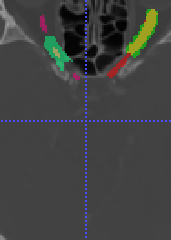}
		\end{minipage}
		\begin{minipage}{0.23\linewidth}
			\includegraphics[width=\textwidth]{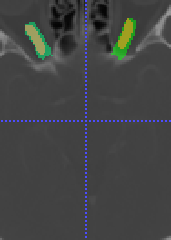}
		\end{minipage}\\
		
		\begin{minipage}{0.23\linewidth}
			\includegraphics[width=\textwidth]{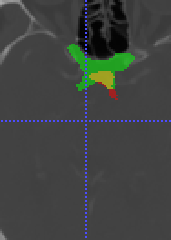}
		\end{minipage}
		\begin{minipage}{0.23\linewidth}
			\includegraphics[width=\textwidth]{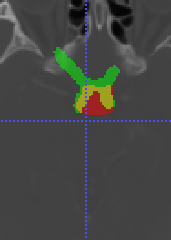}
		\end{minipage}
		\begin{minipage}{0.23\linewidth}
			\includegraphics[width=\textwidth]{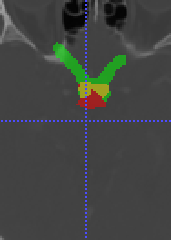}
		\end{minipage}
		\begin{minipage}{0.23\linewidth}
			\includegraphics[width=\textwidth]{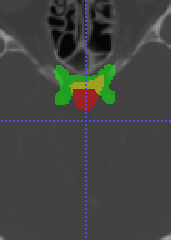}
		\end{minipage}
	\end{center}
\vspace{-0.2in}
	\caption[vis2_2] 
	{ \label{fig:vis2_2} 
		{Visualizations for the rest five anatomies on the first four holdout CT images.}}
\end{figure}

\vspace{-0.1in}
To check the generalization ability of the trained model, we also visualize the segmentation results of the trained model on a small internal dataset in Fig. \ref{fig:vis2} and Fig. \ref{fig:vis2_2}. Visual inspection suggests that the trained model performed well on this independent test set. In general, the performances on larger anatomies are better than small ones (such as optic chiasm), which can be attributed by both manual annotation inconsistencies and algorithmic challenges in segmenting these small regions.

\section{DISCUSSION}\label{sec:discus}
\vspace{-0.2in}
\subsection{Impacts of training datasets}\label{sec:datasetcomp} 
\vspace{-0.1in}
The training datasets we collected come from various sources with annotations done by different groups of physicians with different guiding criteria. It is unclear how the different datasets might contribute the model performance.  For this purpose, we carried out an experiment to test the model performance trained with two different datasets: a) using only the training data provided in the MICCAI head and neck segmentation challenge 2015 (DATASET 1, 38 samples), and b) the combined training data with 216 samples (DATASET 1-3 combined).  In terms of annotations, the first dataset is more consistent with the test dataset, therefore less likely to suffer from annotational inconsistencies.   However, on the other hand, the size of the dataset is much smaller, posing challenges to training deep learning models.


Table \ref{tab:dataset} shows the test performances of a 3D Res U-Net model trained with the above-mentioned two datasets after applying the same training procedure of minimizing Dice loss. We notice a few observations. First, overall the model trained with the larger dataset (DATATSET 1-3) achieves better performance with a 2.5\% improvement over the smaller dataset, suggesting that the larger sample size does lead to a better performance.  Second, although the larger dataset improves performances on average, there are some OARs on which the smaller dataset actually does better, most noticeably, on mandible and optic nerves.  This suggests that there are indeed significant data annotation inconsistencies between different datasets, whose impact on model performance cannot be neglected.  Third, to further check the generalization ability of the model trained with DATASET 1 only, we checked its performance on DATASETS 2-3 and found its performance was generally poor. Altogether, this suggests both annotation quality and data size are important for training deep learning models. How to address inconsistencies in existing datasets is an interesting open question to be addressed in the future.

\begin{table}[ht]
\caption{Performance comparisons of models trained with different datasets.} 
\label{tab:dataset}
\vspace{-0.1in}
\begin{center}       
\begin{tabular}{|c|c|c|} 
\hline
\rule[-1ex]{0pt}{3.5ex}  Datasets &DATASET 1& DATASET 1,2,3\\
\hline
\rule[-1ex]{0pt}{3.5ex}  Brain Stem &58.60&85.91\\
\hline
\rule[-1ex]{0pt}{3.5ex}  Chiasm &39.93&53.26  \\
\hline
\rule[-1ex]{0pt}{3.5ex}  Mandible &94.16&90.59 \\
\hline
\rule[-1ex]{0pt}{3.5ex}  Opt Ner L &74.62&69.80 \\
\hline
\rule[-1ex]{0pt}{3.5ex}  Opt Ner R &73.77&67.50 \\
\hline
\rule[-1ex]{0pt}{3.5ex}  Parotid L &88.83&87.84 \\
\hline
\rule[-1ex]{0pt}{3.5ex}  Parotid R &87.24&86.15 \\
\hline
\rule[-1ex]{0pt}{3.5ex}  Submand. L &78.56&79.91 \\
\hline
\rule[-1ex]{0pt}{3.5ex}  Submand. R &81.83&80.24 \\
\hline
\rule[-1ex]{0pt}{3.5ex}  Average  &75.28&77.91 \\
\hline
\end{tabular}
\end{center}
\end{table}

\vspace{-0.2in}
\subsection{Limitations}\label{sec:futurework}
\vspace{-0.1in}
There are a couple of limitations in the current implementation of AnatomyNet. First, AnatomyNet treats voxels equally in the loss function and network structure. As a result, it cannot model the shape prior and connectivity patterns effectively. The translation and rotation invariance of convolution are great for learning appearance features, but suffer from the loss of spatial information.  For example, the AnatomyNet sometimes misclassifies a small background region into OARs (Fig.\ \ref{fig:vis1},\ref{fig:vis1_2}). The mis-classification results in a partial anatomical structures, which can be easily excluded if the overall shape information can also be learned. A network with multi-resolution outputs from different levels of decoders, or deeper layers with bigger local receptive fields should  help alleviate this issue. 

\begin{table}[ht]
	\caption{Average 95th percentile Hausdorff distance (unit: mm) comparisons with state-of-the-art methods} 
	\label{tab:anatomynethd95}
	\vspace{-0.1in}
	\begin{center}    
		\resizebox{\columnwidth}{!}{   
		\begin{tabular}{|c|c|c|c|} 
			\hline
			\rule[-1ex]{0pt}{3.5ex}  Anatomy & \begin{tabular}{@{}c@{}}\textbf{MICCAI 2015} \\ Range \cite{raudaschl2017evaluation}\end{tabular}  & Ren et al. 2018 \cite{ren2018interleaved} & AnatomyNet\\
			\hline
			\rule[-1ex]{0pt}{3.5ex}  Brain Stem & 4-6 & N/A & 6.42$\pm$2.38 \\
			\hline
			\rule[-1ex]{0pt}{3.5ex} Chiasm & 3-4 & 2.81$\pm$1.56 & 5.76$\pm$2.49 \\
			\hline
			\rule[-1ex]{0pt}{3.5ex} Mand. & 2-13 & N/A & 6.28$\pm$2.21 \\
			\hline
			\rule[-1ex]{0pt}{3.5ex} Optic Ner L & 3-8 & 2.33$\pm$0.84 & 4.85$\pm$2.32 \\
			\hline
			\rule[-1ex]{0pt}{3.5ex} Optic Ner R & 3-8 & 2.13$\pm$0.96 & 4.77$\pm$4.27 \\
			\hline
			\rule[-1ex]{0pt}{3.5ex} Paro. L & 5-8 & N/A & 9.31$\pm$3.32 \\
			\hline
			\rule[-1ex]{0pt}{3.5ex} Paro. R & 5-8 & N/A & 10.08$\pm$5.09 \\
			\hline
			\rule[-1ex]{0pt}{3.5ex} Subm. L & 4-9 & N/A & 7.01$\pm$4.44 \\
			\hline
			\rule[-1ex]{0pt}{3.5ex} Subm. R & 4-9 & N/A & 6.02$\pm$1.78 \\
			\hline
		\end{tabular}
	}
	\end{center}
\end{table}

Second, our evaluation of the segmentation performance is primarily based on the Dice coefficient. Although it is a common metric used in image segmentation, it may not be the most relevant one for clinical applications. Identifying a new metric in consultation with the physicians practicing in the field would be an important next step in order for real clinical applications of the method.  Along this direction, we quantitatively evaluated the geometric surface distance by calculating the average 95th percentile Hausdorff distance (unit: mm, detailed formulation in \cite{raudaschl2017evaluation}) (Table \ref{tab:anatomynethd95}). We should note that this metric imposes more challenges to AnatomyNet than other methods operating on local patches (such as the method by Ren et al. \cite{ren2018interleaved}), because AnatomyNet operates on whole-volume slices and a small outlier prediction outside the normal range of OARs can lead to drastically bigger Hausdorff distance. Nonetheless,  AnatomyNet is roughly within the range of the best MICCAI 2015 challenge results on six out of nine anatomies \cite{raudaschl2017evaluation}. Its performance on this metric can be improved by considering surface and shape priors into the model as discussed above \cite{nikolov2018deep,zhu2018adversarial}.

\section{CONCLUSION}\label{sec:conclu}
In summary, we have proposed an end-to-end atlas-free and fully automated deep learning model for anatomy segmentation from head and neck CT images.  We propose a number of techniques to improve model performance and facilitate model training. To alleviate highly imbalanced challenge for small-volumed organ segmentation, a hybrid loss with class-level loss (dice loss) and focal loss (forcing model to learn not-well-predicted voxels better) is employed to train the network, and one single down-sampling layer is used in the encoder. To handle missing annotations, masked and weighted loss is implemented for accurate and balanced weights updating. The 3D SE block is designed in the U-Net to learn effective features.  Our experiments demonstrate that our model provides new state-of-the-art results on head and neck OARs segmentation, outperforming previous models by 3.3\%.  It is significantly faster, requiring only a fraction of a second to segment nine anatomies from a head and neck CT.  In addition, the model is able to process a whole-volume CT and delineate all OARs in one pass. All together, our work suggests that deep learning offers a flexible and efficient framework for delineating OARs from CT images. With additional training data and improved annotations, it would be possible to further improve the quality of auto-segmentation, bringing it closer to real clinical practice.

\begin{acknowledgments}
We would like to acknowledge the support received from NVIDIA on GPU computing, and helpful discussions with Tang H and Yang L. 
\end{acknowledgments}

\bibliographystyle{IEEEtran} 
\bibliography{aapmtemplate}

\end{document}